\documentclass[10pt,twocolumn,letterpaper]{article}

\usepackage[pagenumbers]{cvpr} 

\definecolor{cvprblue}{rgb}{0.21,0.49,0.74}
\usepackage[pagebackref,breaklinks,colorlinks,allcolors=cvprblue]{hyperref}
\usepackage{multirow}
\usepackage{makecell}
\usepackage{float}

\def\paperID{11517} 
\def\confName{CVPR}
\def\confYear{2025}

\title{CustAny: Customizing Anything from A Single Example}

\author{%
Lingjie Kong$^{1}$\thanks{Equal contribution.}
  Kai Wu$^{2*}$
  Xiaobin Hu$^2$
  Wenhui Han$^{2}$ 
  Jinlong Peng$^2$ \\
  Chengming Xu$^{2}$ 
  Donghao Luo$^{2}$
  Jiangning Zhang$^2$
  Chengjie Wang$^2$
  Yanwei Fu$^{1}$\thanks{Corresponding Author.} \\
  \textsuperscript{1}Fudan University, Shanghai, China~~~
  \textsuperscript{2}Tencent Youtu Lab, Shanghai, China~~~\\
  \texttt{\url{https://lingjiekong-fdu.github.io}}
}

\begin{document}
\maketitle



\begin{abstract}
Recent advances in diffusion-based text-to-image models have simplified creating high-fidelity images, but preserving the identity (ID) of specific elements, like a personal dog, is still challenging.
Object customization, using reference images and textual descriptions, is key to addressing this issue. 
Current object customization methods are either object-specific, requiring extensive fine-tuning, or object-agnostic, offering zero-shot customization but limited to specialized domains. 
The primary issue of promoting zero-shot object customization from specific domains to the general domain is to establish a large-scale general ID dataset for model pre-training, which is time-consuming and labor-intensive. In this paper, we propose a novel pipeline to construct a large dataset of general objects and build the Multi-Category ID-Consistent (MC-IDC) dataset, featuring 315k text-image samples across 10k categories. With the help of MC-IDC, we introduce Customizing Anything (CustAny), a zero-shot framework that maintains ID fidelity and supports flexible text editing for general objects. CustAny features three key components: a general ID extraction module, a dual-level ID injection module, and an ID-aware decoupling module, allowing it to customize any object from a single reference image and text prompt. Experiments demonstrate that CustAny outperforms existing methods in both general object customization and specialized domains like human customization and virtual try-on. Our contributions include a large-scale dataset, the CustAny framework and novel ID processing to advance this field. Code and dataset will be released soon in \url{https://github.com/LingjieKong-fdu/CustAny}.
\end{abstract}

\section{Introduction}

Recent advancements in diffusion-based text-to-image generative models (e.g., SDXL, Improving GAN, PixArt Alpha)~\cite{podell2023sdxl, betker2023improving, chen2023pixartalpha} have empowered users to generate highly realistic images with minimal technical knowledge. However, generating images of specific objects~\cite{chen2023anydoor, wang2024instantid, li2023photomaker, ye2023ip}, such as personal pets, remains challenging due to the difficulty of preserving their unique identity. Object customization, a technique that utilizes reference images and textual descriptions to generate images of specific objects, has emerged as a critical area of research. Existing methods~\cite{ruiz2023dreambooth,zhang2023controlnet,peng2023portraitbooth,pei2024deepfake,wang2024instantid} for object customization, such as DreamBooth, PortraitBooth and InstantID, have limitations in general-purpose object customization: Object-specific models~\cite{ruiz2023dreambooth,chen2023disenbooth}, while powerful, are often inefficient due to the need for extensive fine-tuning for each new object. Conversely, object-agnostic models~\cite{li2023photomaker,wang2024instantid} have limited capabilities and often struggle with diverse object types.

Unlike existing customizing applications tailored to one specific task like face generation or virtual try-ons, this paper focuses on general object customization from a single example, as shown in Fig.~\ref{fig:intro}. Specifically, it addresses zero-shot object customization in various scenarios where no fine-tuning on specific objects is available. Achieving this first requires a large-scale dataset containing various categories of objects for model pre-training. At present, however, there are few publicly released datasets in the field of zero-shot customization for general objects. To address this issue, we propose a novel construction pipeline for general ID dataset and introduce the Multi-Category ID-Consistent (MC-IDC) dataset, the first large-scale dataset for zero-shot object customization across diverse scenarios. It includes over 315K high-quality images across 10K categories, such as human faces, animals, clothing, and tools, supporting a wide range of object types and domains. The dataset features reference-target image pairs with consistent object IDs, including segmentation masks and text captions, making it ideal for ID-consistent generation tasks and advancing research in general object customization.

Based on our  MC-IDC, we design a zero-shot general object customization framework which is able to Customize Anything (CustAny) from a single example. The innovation of our framework is mainly reflected in managing and utilizing identity information beyond the narrow focus of previous work. Specifically, the CustAny framework follows three essential steps: ID extraction, ID injection, and ID disentanglement. These steps address three critical questions:
(1) How to capture enough identity details to handle a variety of complex customization tasks?
(2) How to add this identity information to the model while still enabling text-based edits?
(3) How to reduce the impact of irrelevant features from the reference object on the final results?

Formally, CustAny features an ID extraction module, dual-level ID injection, and an ID-aware decoupling module, offering innovative ID handling methods.
Unlike prior methods~\cite{li2023photomaker,ye2023ip,wang2024instantid} that use a single model for object representations, we leverage multiple pre-trained self-supervised models to capture more detailed ID-related features for high-ID-fidelity customization. To integrate this information into the diffusion model, we use a dual-level ID injection strategy: globally, by merging semantic-level ID features with text descriptions, and locally, by injecting patch tokens through cross-attention modules, preserving both ID fidelity and text editability.

Furthermore, we identify that ID information is often mixed with redundant non-ID elements, like object motion and orientation, which can disrupt both text editing and ID retention. To solve this, we introduce an ID-aware decoupling module during training to help the model separate ID-related properties from non-ID elements, enhancing both ID fidelity and generation diversity. With these advanced ID processing modules, CustAny enables zero-shot customization for various objects while supporting flexible text-based editing. As shown in Fig.~\ref{fig:intro}, CustAny applies to object customization in various scenarios such as human customization, cartoon character personalization, virtual try-on.
We conduct extensive experiments showing that CustAny achieves state-of-the-art performance in general object customization and outperforms task-specific methods in areas like human customization and virtual try-on.

In summary, we have these contributions:
(1) \textbf{Dataset}: We propose a novel construction pipeline for general ID dataset, and present a large-scale dataset, MC-IDC, with 315K images across 10K categories for zero-shot object customization.
(2) \textbf{CustAny Framework}: We design a zero-shot framework for general object customization in text-to-image generation, focusing on ID fidelity and text-based editing.
(3) \textbf{ID Processing}: CustAny uses multiple self-supervised models and a dual-level ID injection strategy to integrate ID information while preserving text-editability. We present  a module to separate ID from non-ID elements, improving ID fidelity and generation diversity.
(4) \textbf{Good Results}: CustAny outperforms task-specific methods in general and specialized object customization tasks, including human customization and virtual try-on.

\begin{figure*}
    \centering
    \vspace{-0.1in}
    \includegraphics[width=0.75\linewidth]{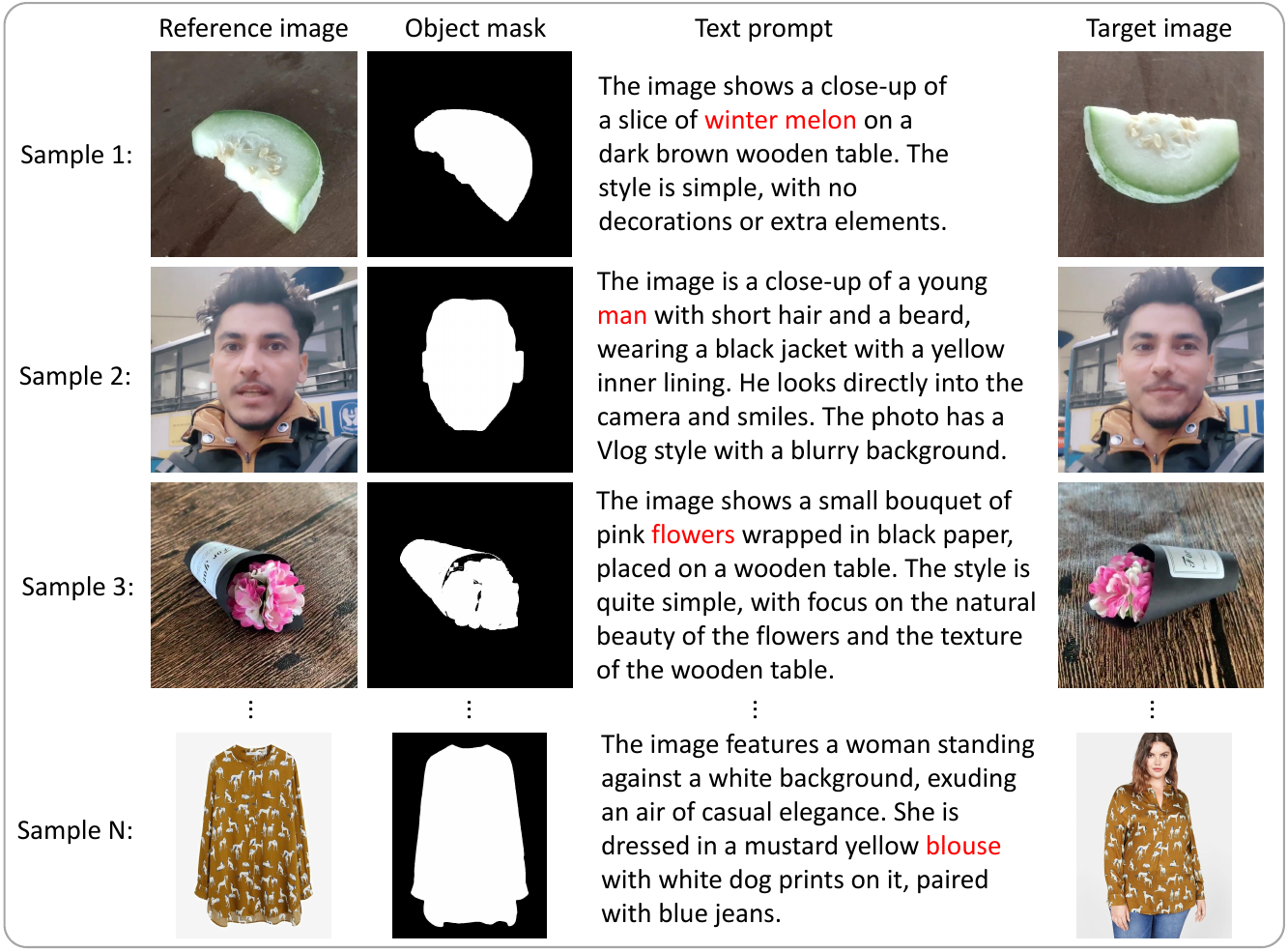}
    \vspace{-0.15in}
    \caption{Illustration of our general ID dataset, MC-IDC. In each sample, the reference image with the object mask provides ID information, the text prompt offers semantic-level guidance for generation, and the target image serves as the ground truth.     \label{fig:fulu-dataset}}
\vspace{-0.15in}
\end{figure*}

\section{Related Work}

\noindent \textbf{Diffusion models}.
Diffusion models have demonstrated their effectiveness for text-to-image generations.
The DDPM model pioneered in this field, using a diffusion and denoising process to create a mapping between Gaussian and image distributions. The Latent Diffusion Model (LDM)~\cite{rombach2022LDM} took this a step further by applying the diffusion model to a latent space rather than pixel space, leading to the creation of text-to-image diffusion models like Stable Diffusion (SD), Midjourney, and DALLE-3~\cite{ramesh2022dalle}. One main research problem for the diffusion model is its structure. Other than the original UNet, DiT~\cite{peebles2023dit} and Pixart-$\alpha$~\cite{chen2023pixartalpha} adopt transformer as the backbone structure. Other researchers have also focused on enhancing image fidelity regarding complicated concepts. 
By manipulating the cross-attention maps according to text prompts, ~\cite{chefer2021transformerbeyond} ensures that required objects are sufficiently generated in the images. 
~\cite{ge2023expressive} proposed to process the text prompts to decompose all target attributes, which can be further used to guide the denoising procedure with the help of different objective functions. 
To enhance content consistency, Tewel et al.~\cite{tewel2022zerocap} adopted an inner localization method by calculating the cross attention. 
Our method is generally built on diffusion models. While different from these basic models for plain image generations, we focus on the more challenging topic of zero-shot any object customization. 

\noindent  \textbf{Image customization}. 
Object customization techniques generally fall into two categories: object-specific and object-agnostic (zero-shot) methods. Object-specific approaches, like DreamBooth~\cite{ruiz2023dreambooth} and DisenBooth~\cite{chen2023disenbooth}, fine-tune pretrained diffusion models by incorporating new, object-specific identifiers from a small set of reference images. While effective, these methods require extensive fine-tuning for each new object, which limits their scalability and efficiency.  In contrast, object-agnostic or zero-shot approaches such as PhotoMaker~\cite{li2023photomaker} and InstantID~\cite{wang2024instantid}, leverage large-scale training datasets to customize images without the need for additional fine-tuning. Although more efficient, these methods are often domain-specific, focusing on tasks like human customization, facial manipulation, or virtual try-on applications. For example, PhotoMaker trains a person ID extraction network, while InstantID applies ControlNet~\cite{zhang2023controlnet} to control facial expressions based on pretrained embeddings, and MagicCloth~\cite{chen2024magiccloth} uses ControlNet to customize garment details for virtual try-on.
These existing methods, however, have limitations when it comes to general-purpose object customization. Object-specific models are powerful but inefficient, requiring time-intensive training for each new item, while object-agnostic models tend to be narrow in focus and are often limited to specific types of objects or applications.
In this work, we aim to bridge these gaps by combining the versatility of object-specific methods with the efficiency of zero-shot methods. As illustrated in Fig.~\ref{fig:intro}, our approach targets the ability to customize any object based on a single example image and text description, achieving generalization across diverse categories with minimal fine-tuning. This design marks a significant step toward general object customization across various domains, an area that remains largely unexplored.

\begin{figure*}
    \centering
    \includegraphics[width=0.82\textwidth]{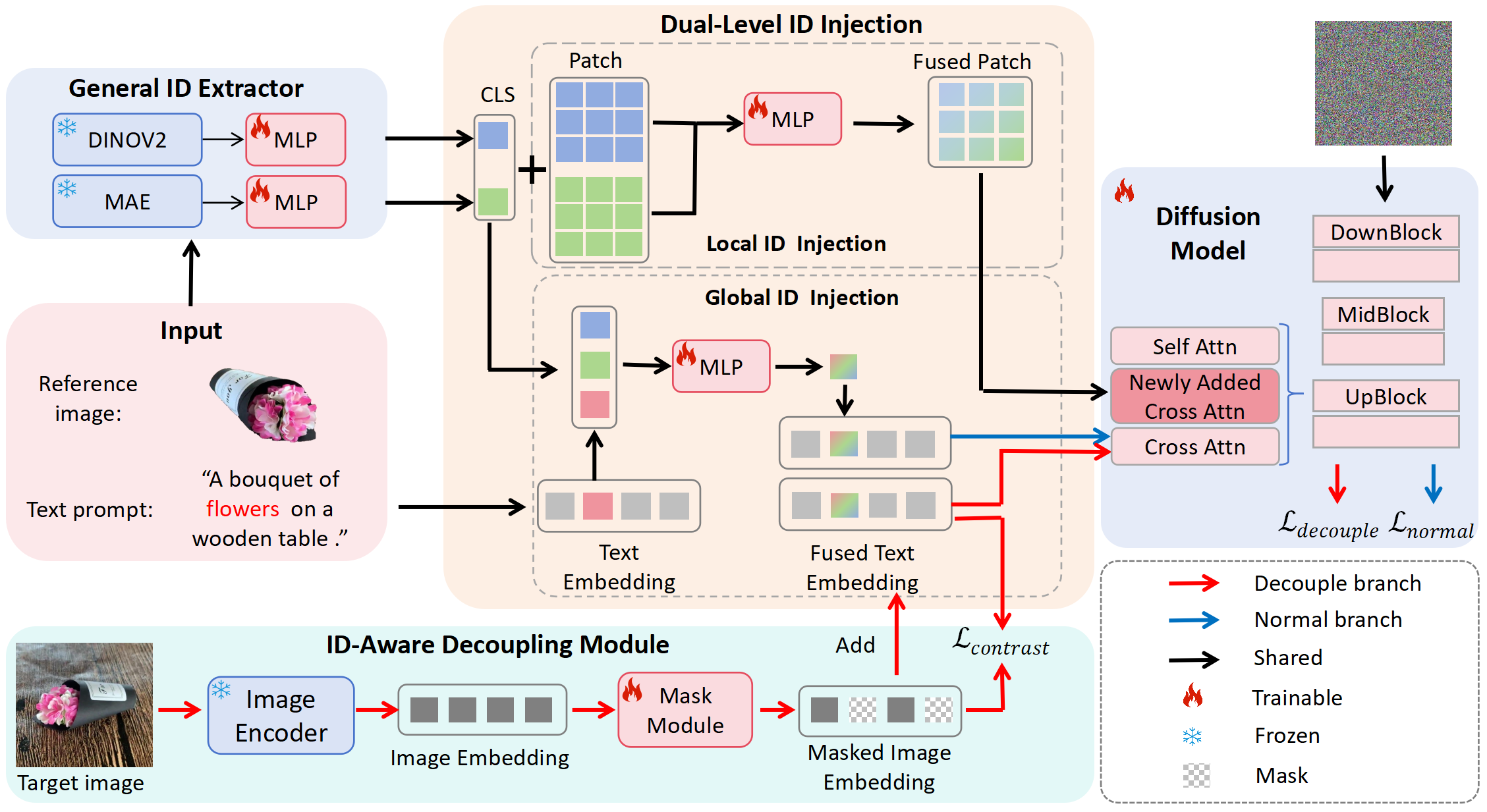}
    \vspace{-0.1in}
    \caption{Overview of our CustAny. CustAny is a zero-shot text-to-image customization method for general objects, consisting of general ID extractor, global-local dual-level ID injection, and ID-aware decoupling module.}
    \label{fig:overview}
      \vspace{-0.2in}
\end{figure*}

\section{MC-IDC Dataset}
\label{sec:dataset}

We curate the Multi-Category ID-Consistent (MC-IDC) dataset, the first large-scale dataset for general object customization research. We present the overall illustration of the MC-IDC as shown in Fig.~\ref{fig:fulu-dataset}. The dataset's merits and construction process are described below, with additional details in  Supplementary.

\noindent \textbf{Construction pipeline}. The MC-IDC  is constructed by the following four steps:
(1) \textit{Data collection}.
MC-IDC dataset contains diverse sources including web-crawled images, movies and publicly available datasets. To facilitate the subsequent generation of image pairs, most of the data we collect are video datas or multi-view datas. To ensure the high quality of our dataset, we delete the images with a resolution less than 300$\times$300.
\noindent (2) \textit{Instance detection and segmentation}.
For some publicly available datasets or web-crawled images which do not provide segmentation annotations, we use  advanced instance segmentation model~\cite{wang2023internimage} for instance detection and segmentation. For movie dataset, we extract frames with 10fps and do object tracking with~\cite{peng2020chaintracker}, aiming to establish the ID connection across frames. 
\noindent (3) \textit{Image pair generation}.
For most data samples which are from a clip of video or a group of multi-view data, we randomly select two frames containing the same object and perform a random crop around the object to form the reference-target image pair. For the other data samples from the single image dataset, we perform random data augmentations on the single image two times, and take them as the reference image and the target image respectively.
\noindent (4) \textit{Text prompt generation}.
In each image pair, the target image is densely captioned with the state-of-the-art large vision language model~\cite{bai2023qwen}, serving as the text prompt to guide the generation.


\noindent \textbf{Merits of MC-IDC}:
Our MC-IDC dataset contains 315k high-quality samples across 10k+ categories and diverse domains, with each sample featuring reference-target image pairs, segmentation masks, and text captions to support ID-preserving, diverse object customization. Particularly, it has these merits: 
\noindent (1)\textit{Diverse domains and categories:}
Our MC-IDC dataset contains around 315k samples across 10k+ categories, including human faces, animals, clothes, and tools. It features images from diverse domains, such as real-world photos, animations, model-generated content, and movies, enabling general ID understanding. Further details are provided in the Supplementary.
\noindent (2)\textit{High quality:}
With an average resolution of 1039$\times$950, our dataset ensures high-quality training and inference for our model.
\noindent (3) \textit{Reference-target image pair:}
Each sample includes a reference image with a segmentation mask, a target image, and a text caption. The reference image provides ID information, while the target image and text guide semantic generation. The reference and target images depict the same object in different states, sharing the same ID and varying non-ID elements like motion and direction. This setup ensures ID preservation while maintaining generation diversity.

\section{Method}

\noindent \textbf{Overview}.
Given a reference image and text prompt, our goal is to generate an image that retains the identity (ID) of the object in the reference while modifying non-ID elements, like motions and backgrounds, based on the prompt. Thus we introduce CustAny, a zero-shot text-to-image customization framework for general objects, as  in Fig.~\ref{fig:overview}. 
We first use an ID extractor to capture ID information from the reference image with a segmentation mask. Dual-level ID injection is then applied to embed the ID into the diffusion model globally and locally, preserving text-editing capabilities. Finally, an ID-aware decoupling module separates ID details from non-ID elements, improving ID fidelity and text-editing accuracy.

\subsection{General ID Extraction}
\label{sec:id_extract}

To extract comprehensive ID information, we use a combination of self-supervised models, DINOv2~\cite{oquab2024dinov2} and MAE~\cite{he2021masked}, to extract representations for general objects in the reference image. Specifically, previous methods, such as~\cite{li2023photomaker, wang2024instantid}, which target specific domains like faces, typically use a single model like CLIP~\cite{radford2021learning} to extract object features. However, we find that using only one pre-trained model—whether CLIP, DINOv2, or MAE—leaves gaps in capturing features necessary for high-ID-fidelity in general objects (see Fig.~\ref{fig:abladino} and Tab.~\ref{tab:extraction}). Essentially, while CLIP as in~\cite{chen2023anydoor,wang2024instantid} lacks sufficient detail, DINOv2 (despite its strong details due to contrastive learning) is color-insensitive because it is trained with color-augmented techniques like ColorJitter. MAE, however, retains color sensitivity due to its reconstruction-based training. Thus, combining DINOv2 and MAE allows us to capture both detail and color for more accurate ID extraction.

We combine DINOv2 and MAE, which complement each other, to capture comprehensive information for the challenging task of customizing diverse general objects. To align and prepare these features for injection, we use a two-layer MLP. The extraction process for general object representations is as follows,
\begin{equation}
   f_{dino}^{C}, f_{dino}^{P} = MLP(F_{dino}(I_{ref}\odot M_{ref})),
\end{equation}
\begin{equation}
   f_{mae}^{C}, f_{mae}^{P} = MLP(F_{mae}(I_{ref}\odot M_{ref})),
\end{equation}
where \(\odot\) represents element-wise multiplication; and we have the class tokens and patch tokens \(f_{dino}^{C}, f_{dino}^{P}\), \(f_{mae}^{C}, f_{mae}^{P} \) extracted by DINOv2 and MAE respectively. $F_{dino}$ and $F_{mae}$ are the DINOv2 and MAE backbone. $I_{ref}$ denotes the reference image; and $M_{ref}$ is the segmentation mask of the interested object in the reference image.
Consequently, tokens derived from our general ID extractor not only offer intricate details from DINOv2, but also have sufficient colour and structural information from MAE.




\subsection{Dual-level ID Injection}
\label{sec:injection}

We aim to inject maximum extracted ID information into the diffusion UNet without compromising its text-editing ability. Our dual-level ID injection mechanism separately embeds global semantic ID and local fine-grained ID into the UNet, ensuring flexible text edits and high ID fidelity in customizing general objects.

\noindent \textbf{Global ID injection}. \label{ssec_global}
In text-to-image generation, text prompts guide diffusion models to create diverse images. However, diffusion models cannot reliably generate images with a specific object ID based on text alone. To embed object ID information into the text while preserving editability for other elements like background and motion, we introduce a global ID injection mechanism. Specifically, we merge the semantic-level class tokens $(f_{dino}^C, f_{mae}^C)$ with the class word in the text embedding, such as ``dog'' or ``cat", 
\begin{equation}
   f_{fuse}^{C} = MLP(Concat(f_{text}^{C}, f_{dino}^{C}, f_{mae}^{C})),
\label{eq_fuse_c}
\end{equation}
%
%
where \(f_{fuse}^{C}\) is  fused class token and \(f_{text}^{C}\) is  class word embedding. We insert \(f_{fuse}^{C}\) into the class word position in the text embeddings, obtaining the global-level condition \(c_g\) with ID-related information, as  in Fig.~\ref{fig:overview}. The global-level condition \(c_g\) interacts with  cross-attention module of UNet, similar to standard text-to-image models~\cite{rombach2022LDM} as,
\begin{equation}
   Z_g = Attention(Q_g,K_g,V_g),
\end{equation}
where $Q_g=Z_gW_q$, $K_g=c_{g}W_k$, $V_g=c_{g}W_v$ are the query, key, and value of cross-attention module, in which $W_q$, $W_k$, $W_v$ are the weight matrices of the trainable linear projection layers, and $Z_g$ is the query feature containing the latent input information.



\noindent \textbf{Local ID injection}. 
The global-level condition \(c_g\) contains limited object details, making it insufficient for complex customization tasks. To provide more ID-related information, we propose the local ID injection mechanism, where we fuse the extracted patch tokens \(f_{dino}^P\) and \(f_{mae}^P\) using a two-layer MLP to create a unified representation \(c_l\),
\begin{equation}
   c_l = MLP(f_{dino}^P) + MLP(f_{mae}^P),
\end{equation}
where \(c_l\) serves as the local-level condition for the diffusion UNet. Then, we add one cross-attention module in each upblock of the diffusion UNet in order to inject the ID-related details into the model without compressing $c_l$ dimensions, feeding the model with as much ID information as possible.
\begin{equation}
   Z_l = Attention(Q_l,K_l,V_l),
\end{equation}
where $Q_l=Z_lW_q$, $K_l=c_{l}W_k$, $V_l=c_{l}W_v$ are the query, key, and value of cross-attention module, in which $W_l$, $W_l$, $W_l$ are the weight matrices of the trainable linear projection layers, and $Z_l$ is  query feature containing the latent input information.
By dual-level ID injection, we have the model with sufficient ID-related information at both  semantic and detail levels, without impairing its text-editing ability.

\subsection{Training with ID-Aware Decoupling}
\label{sec:decouple}


The ID-related information injected into the diffusion model is often mixed with non-ID details like motion, direction, and size. Without proper guidance, the model may learn both together, disrupting text editing and limiting generation diversity. For example, in a reference image of ``a standing person'' the posture ``standing'' could be injected, preventing the generation of ``a sitting person" from a text prompt, leading to less diverse outputs. 
To address this, we propose an ID-aware decoupling module, inspired by previous works~\cite{chen2023disenbooth,he2021masked}, which includes a decoupling branch and associated losses during training to help the model distinguish ID information from other features.

\noindent \textbf{Decoupling branch and normal branch}.
In the decoupling branch, we first extract the target image features $f_{tar}$ through CLIP~\cite{radford2021learning} to make an image embedding prior~\cite{ramesh2022dalle}. Next, we mask out the ID information contained in the image embedding by a trainable feature mask \(m_{id}\), and thus the masked image feature $f_{msk}=f_{tar}\odot m_{id}$ contains only non-ID information of the target image, where \(\odot\) represents element-wise multiplication.
Then, we add the masked image feature $f_{msk}$ to the fused class token \(f_{fuse}^{C}\) and then feed it to the diffusion UNet for generation following the injection way mentioned above. Yet in the normal branch, we inject \(f_{fuse}^{C}\) to the model without $f_{msk}$. The generation process of two branches can be illustrated as follows,
\begin{equation}
    f_{msk} \oplus f_{fuse}^{C}  \Rightarrow  I_{tar},\hspace{0.2cm} f_{fuse}^{C}  \Rightarrow  I_{tar},\hspace{0.2cm} f_{msk}\perp f_{fuse}^{C},
\label{eq:newdecoupleprocess}
\end{equation}
where \(\oplus\) denotes element-wise addition, \(I_{tar}\) for target image, and \(\perp\) indicates that the features on both sides are independent. Through training in both branches, we push non-ID information into  masked image feature \(f_{msk}\) and retain only ID-related information in the fused class token \(f_{fuse}^C\).


\noindent \textbf{Training strategy}. We use three losses in our training stage to ensure that the ID-aware decoupling module takes effect. Specifically,
We utilize the following denoising losses to train each branch,
\begin{equation}
\mathcal{L}_{decouple} = \| \epsilon- \epsilon_\theta\big(x_t, c_g,f_{msk},c_l, t\big)\|^2,
\label{eq:newdecoupleloss}
\end{equation}
\begin{equation}
\mathcal{L}_{normal} = \| \epsilon- \epsilon_\theta\big(x_t, c_g,c_l, t\big)\|^2,
\label{eq:newnormal}
\end{equation}
where $t$ is the randomly sampled time step; and $x_t$ represents the noisy latent of the target image; $\epsilon$ is the ground truth noise and $\epsilon_\theta$ is the predicted noise.
$c_g$ and $c_l$ shall be the global-level and local-level conditions acquired in the dual-level ID injection module.
Additionally, we design a contrastive loss, ensuring that the masked image feature $f_{msk}$ captures non-ID information and the fused class token $f_{fuse}^{C}$ remains free of it, as below, 
\begin{equation}
    \mathcal{L}_{contrast} = Sim(f_{fuse}^{C}, f_{msk}),
\end{equation} 
where $Sim$ is instantiated as cosine similarity. In summary, the training loss is the sum of the three loss functions mentioned above,
\begin{equation}
    \mathcal{L} = \alpha_{1}\mathcal{L}_{normal}+\alpha_{2}\mathcal{L}_{decouple}+\alpha_{3}\mathcal{L}_{contrast},
\end{equation} 
where hyperparameters $\alpha_{1},\alpha_{2},\alpha_{3}$ are coefficients which are set to 2.0, 1.0, 0.5 respectively. 





%

\begin{table*} \small
\setlength{\abovecaptionskip}{0cm}
\vspace{-0.3in }
\centering
\tabcolsep=0.4cm
\caption{Quantitative comparison among zero-shot object customization methods in general domain and two specific popular domains namely human customization and virtual try-on.}
\begin{tabular}{clccccccc}
    \toprule
    Domains & Methods & FID$\downarrow$ & CLIP-i$\uparrow$ & CLIP-t$\uparrow$ & DINO-i$\uparrow$ & FaceSim$\uparrow$ & DiverSim-i$\downarrow$  \\ \midrule
    \multirow{2}*{General objects} 
    & IP-Adapter & 70.32 & 77.18 & 28.03 & 44.94 & - & 84.43±0.66  \\
    & Ours & \textbf{47.09} & \textbf{82.16} & \textbf{29.27} & \textbf{65.13} & - & \textbf{74.38±3.99}  \\ \midrule
    \multirow{4}*{\makecell[c]{Human \\ customization}} 
    & IP-Adapter & 102.69 & 72.17 & 29.32 & 41.44 & 65.51 & -  \\
    & PhotoMaker & 106.35 & 71.80 & 32.13 & 44.62 & 64.10 & \textbf{-}  \\
    & InstanceID & 113.18 & 75.87 & \textbf{32.89} & 49.26 & 63.26 & -  \\
    & Ours & \textbf{86.40} & \textbf{79.60} & 30.88 & \textbf{57.44} & \textbf{78.54} & -  \\ \midrule
    \multirow{3}*{Virtual try-on}
    & MagicClothing & 126.09 & 76.53 & 21.40 & 29.10 & - & 89.36±0.40  \\
    & IP-Adapter & 104.47 & 81.99 & \textbf{25.03} & 59.39 & - & 71.28±5.06 \\
    & Ours & \textbf{50.65} & \textbf{83.82} & 22.42 & \textbf{66.24} & - & \textbf{71.27±3.63} \\ \bottomrule
\end{tabular}%
\label{tab:main}
\vspace{-0.1in }
\end{table*}

\section{Experiment}



\noindent \textbf{Implementation details}.
We use SD1.5 as the backbone model for compatibility with open communities like Civitai.com, and apply the multi-scale training mode from~\cite{podell2023sdxl} to handle varied image resolutions. The ID-aware decoupling module uses the CLIP encoder from~\cite{rombach2022stablediffusion}, while MAE-ViT-h/14 and DINOv2-ViT-g/14-reg4 are used to extract ID features. Training involves a 1e-5 learning rate, batch size of 32, and 6 epochs on 32 V100 GPUs, taking about 30 hours. During inference, we perform 50 denoising steps and set the classifier-free guidance scale to 7.



\noindent \textbf{Competitors}.
To demonstrate CustAny's generality, we compare it with both general object customization methods and task-specific methods in human customization and virtual try-on. We compare CustAny with IP-Adapter~\cite{ye2023ip} for general object customization, PhotoMaker~\cite{li2023photomaker} and InstantID~\cite{wang2024instantid} for human customization, and MagicClothing~\cite{chen2024magiccloth} for virtual try-on. All methods are zero-shot and use a single reference image, so we exclude object-specific methods like DreamBooth~\cite{ruiz2023dreambooth} and DisenBooth~\cite{chen2023disenbooth} that require multiple reference images and fine-tuning.



\noindent \textbf{Evaluation dataset}.
Our evaluation dataset consists of 1,000 text-image samples, covering general objects, human data, and virtual try-on data in a 4:3:3 ratio. None of these samples are included in the training set.


\noindent \textbf{Evaluation metrics}.
As PhotoMaker~\cite{li2023photomaker}, we utilize DINO-i~\cite{oquab2024dinov2} and CLIP-i~\cite{radford2021clip} to measure the ID fidelity and use CLIP-t to measure the prompt fidelity. We leverage FID~\cite{heusel2017gans} to assess the generation quality. For human customization, we additionally calculate the face similarity (FaceSim) with FaceNet~\cite{schroff2015facenet} as commonly done in~\cite{li2023photomaker,wang2024instantid}. Further,
we introduce DiverSim-i, a novel metric that measures the average DINO similarity across images generated from diverse text prompts. A lower DiverSim-i value indicates a stronger model ability to generate diverse images matching the prompts. Details are in Supplementary.



\begin{figure*}
    \setlength{\abovecaptionskip}{0cm}
    \centering
    \includegraphics[width=0.8\textwidth]{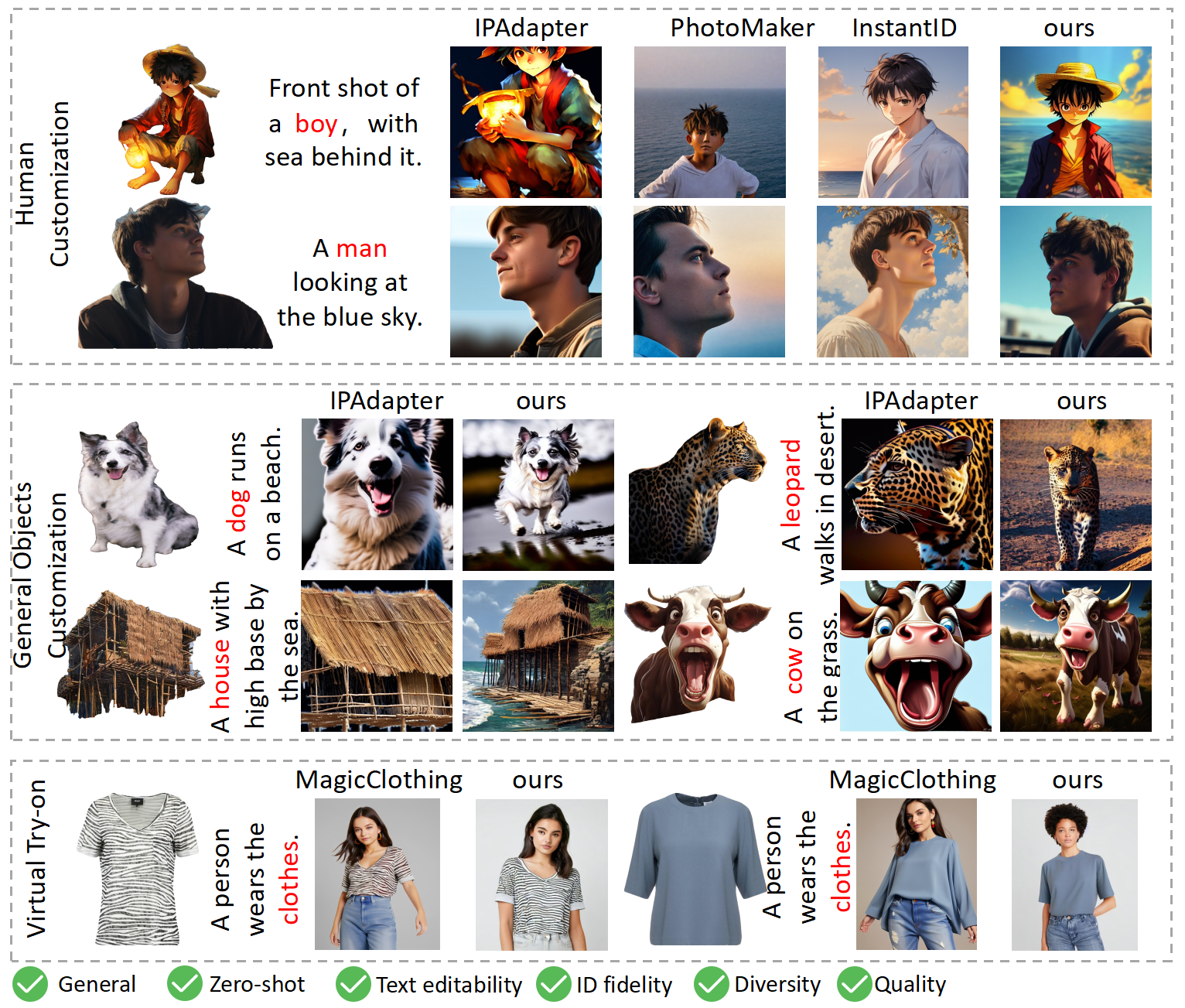}
    \caption{Qualitative results on general domains and two specific specific domains: human customization and virtual try-on.  CustAny 
    exhibits great ID-preserving ability with better text controls and more diverse generations on both general objects and specialized domains. \label{fig:qualitative} }
    \vspace{-0.3in}
\end{figure*}

\begin{figure}
    \centering
    \vspace{-0.2in}
    \includegraphics[width=\linewidth]{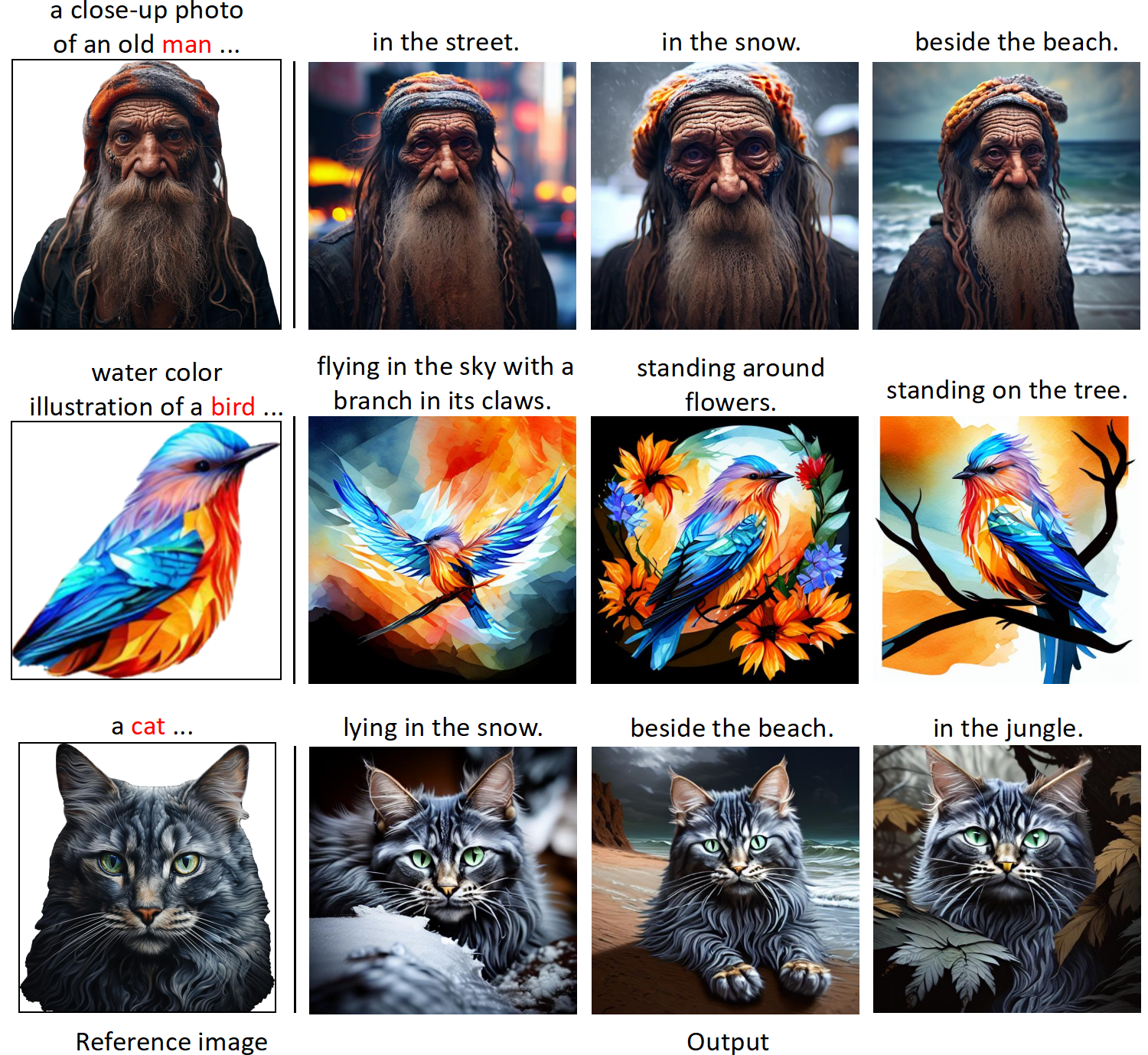}
    \vspace{-0.2in}
    \caption{Results: the same reference in different text prompts.     \label{fig:k2-fulu-genid}}
\vspace{-0.1in}
\end{figure}

\subsection{Quantitative and Qualitative Analysis}
\noindent \textbf{Quantitative results}. Our CustAny outperforms previous works on general object customization for all metrics, and achieve comparable or better performance in contrast with specific methods tailored for specialized tasks such as human customization and virtual try-on, as shown in Tab.~\ref{tab:main}.
Specifically, we achieve higher ID fidelity for CLIP-i and DINO-i, better prompt fidelity for CLIP-t, and higher generation quality evaluated by FID. Furthermore, our method exhibits better diversity on various scenarios as presented by DiverSim-i. 

\begin{table} \small
\setlength{\abovecaptionskip}{0cm}
\centering
\vspace{-0.0in}
\caption{Ablation study on the ID extraction methods.} 
\label{tab:extraction}
\tabcolsep=0.4cm
\begin{tabular}{ccccc}
\toprule
Methods & FID $\downarrow$ & CLIP-i $\uparrow$  & DINO-i $\uparrow$ \\
\midrule
CLIP & 49.11 & 79.58  & 59.45 \\
DINO & 48.89 & 80.82  & 63.71 \\
MAE & 49.00 & 79.09  & 59.79 \\
Ours & \textbf{47.50} & \textbf{81.86}  & \textbf{65.12} \\
\bottomrule
\end{tabular}%
\vspace{-0.2in}
\end{table}

\begin{table} \small
\setlength{\abovecaptionskip}{0cm}
\vspace{-0.1in}
\centering
\caption{Ablation study on the ID injection methods.}
\label{tab:injection}
\tabcolsep=0.4cm
\begin{tabular}{ccccc}
\toprule
Methods & FID $\downarrow$ & CLIP-i $\uparrow$  & DINO-i $\uparrow$ \\
\midrule
Global & 49.78 & 78.76  & 60.67 \\
Local & 48.66 & 81.24  & 62.89 \\
Ours & \textbf{47.50} & \textbf{81.86}  & \textbf{65.12} \\
\bottomrule
\end{tabular}%
\vspace{-0.1in}
\end{table}

\begin{table} \small
\setlength{\abovecaptionskip}{0cm}
\vspace{-0.0in}
\centering
\caption{Ablation study on the ID-aware decoupling module.}
\label{tab:decoupling}
\tabcolsep=0.4cm
\begin{tabular}{ccccc}
\toprule
Methods & FID $\downarrow$ & CLIP-i $\uparrow$  & DINO-i $\uparrow$ \\
\midrule
w.o. decoupling & 47.50 & 81.86  & 65.12 \\
with decoupling & \textbf{47.09} & \textbf{82.16} & \textbf{65.13} \\
\bottomrule
\end{tabular}
\vspace{-0.3in}
\end{table}

\noindent \textbf{Qualitative analysis}.
The CustAny exhibits outstanding capabilities of high-quality customization for general objects, and even beat task-specialized methods in the specific domains, such as human customization and virtual try-on, as shown in Fig.~\ref{fig:qualitative}. We also show the visual results of the same reference picture under different text prompts in Fig.~\ref{fig:k2-fulu-genid}.
Our method is capable of generating high-ID-fidelity images given one single reference image, and simultaneously editing non-ID elements like posture or background in accordance with text prompts, which endows the model with the ability to generate diverse images, such as various scenarios and different postures. Furthermore, benefiting from the ID-aware decoupling module, our model is capable of maintaining the ID fidelity of objects while diversifying the non-ID elements such as motions and directions compared to the reference images, even without corresponding guidance of text prompts. We show more visual results in  Supplementary.
\begin{figure}
    \setlength{\abovecaptionskip}{0cm}
    \centering
    \includegraphics[width=1.0\columnwidth]{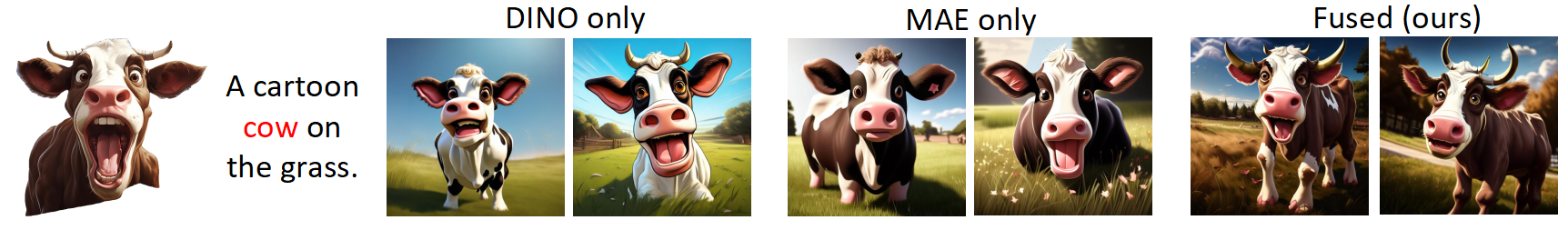}
    \caption{Ablation on different ID extraction methods in the quantitative perspective. Our ID extractor has both the rich details from DINOv2 and the color information from MAE.  \label{fig:abladino} }  
    \vspace{-0.05in}
\end{figure}
\begin{figure}
    \setlength{\abovecaptionskip}{0cm}
    \centering
    \includegraphics[width=0.85\columnwidth]{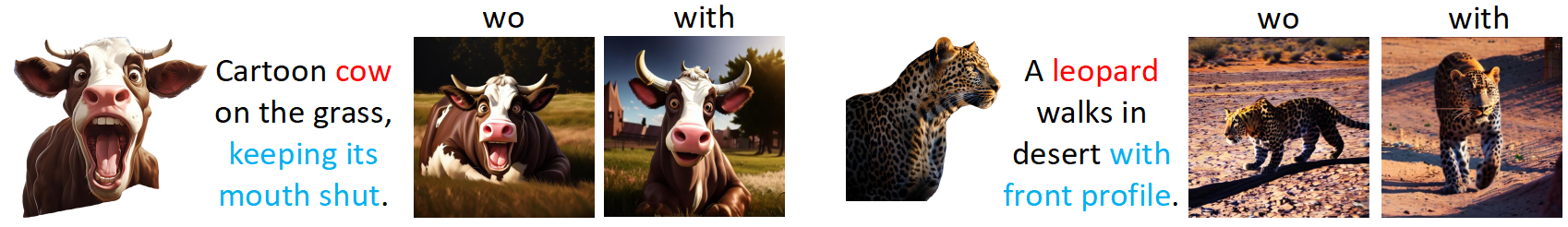}
    \caption{Ablation on whether to add the ID-aware decoupling module in the quantitative perspective. The model with the ID-aware decoupling module demonstrates stronger text control ability, as shown by the blue text.\label{fig:abladisen} }
    
    \vspace{-0.2in}
\end{figure}


\noindent \textbf{Effectiveness of general ID extraction}.
We compare our proposed general ID extractor with DINOv2~\cite{oquab2024dinov2}, MAE~\cite{he2021masked} and CLIP~\cite{radford2021clip}. As in Tab.~\ref{tab:extraction}, our method performs the best, whether in terms of FID, which measures the quality of generation, or in terms of CLIP-i and DINO-i, which measure ID fidelity. As in Fig.~\ref{fig:abladino}, merely using DINOv2 as the extractor fails to maintain color consistency, while solely using MAE as the extractor lacks sufficient details. In contrast, our method combines the advantages of both DINOv2 and MAE, which enables our model to customize images with rich details and the consistent color of the object in reference images.

\noindent \textbf{Benefits of dual-level ID injection}.
We explore the effectiveness of our dual-level ID injection module compared to only global-level injection and only local-level injection separately. As shown in Tab.~\ref{tab:injection}, by combining global and local injection, CustAny achieves the best generation results both in quality as measured by FID and ID-fidelity as measured by CLIP-i and DINO-i. 

\noindent \textbf{Benefits of ID-aware decoupling}.
We conduct a comparative experiment on whether to add the ID decoupling module to the model during training, aiming to verify its effectiveness.
As in Tab.~\ref{tab:decoupling}, the model trained with the ID-aware decoupling module achieves higher ID-fidelity scores in terms of CLIP-i and DINO-i, which means that the decoupling module can help the model to discern ID information embedded in object representations, thereby generating results with better ID consistency. Further, we visualize the results by models trained with and without decoupling in Fig.~\ref{fig:abladisen}. The model trained with decoupling exhibit enhanced capabilities in distinguishing ID information from non-ID elements such as motions (open or shut the mouth) and directions (front-facing or side-facing) of objects of interest. The enhanced discrimination ability allows the model to mitigate the influence of non-ID information during generation, thereby better preserving the text editing capabilities. 

\subsection{Additional Qualitative Applications}

\begin{figure}
\vspace{-0.15in}
    \setlength{\abovecaptionskip}{0cm}
    \centering
    \includegraphics[width=1.0\columnwidth]{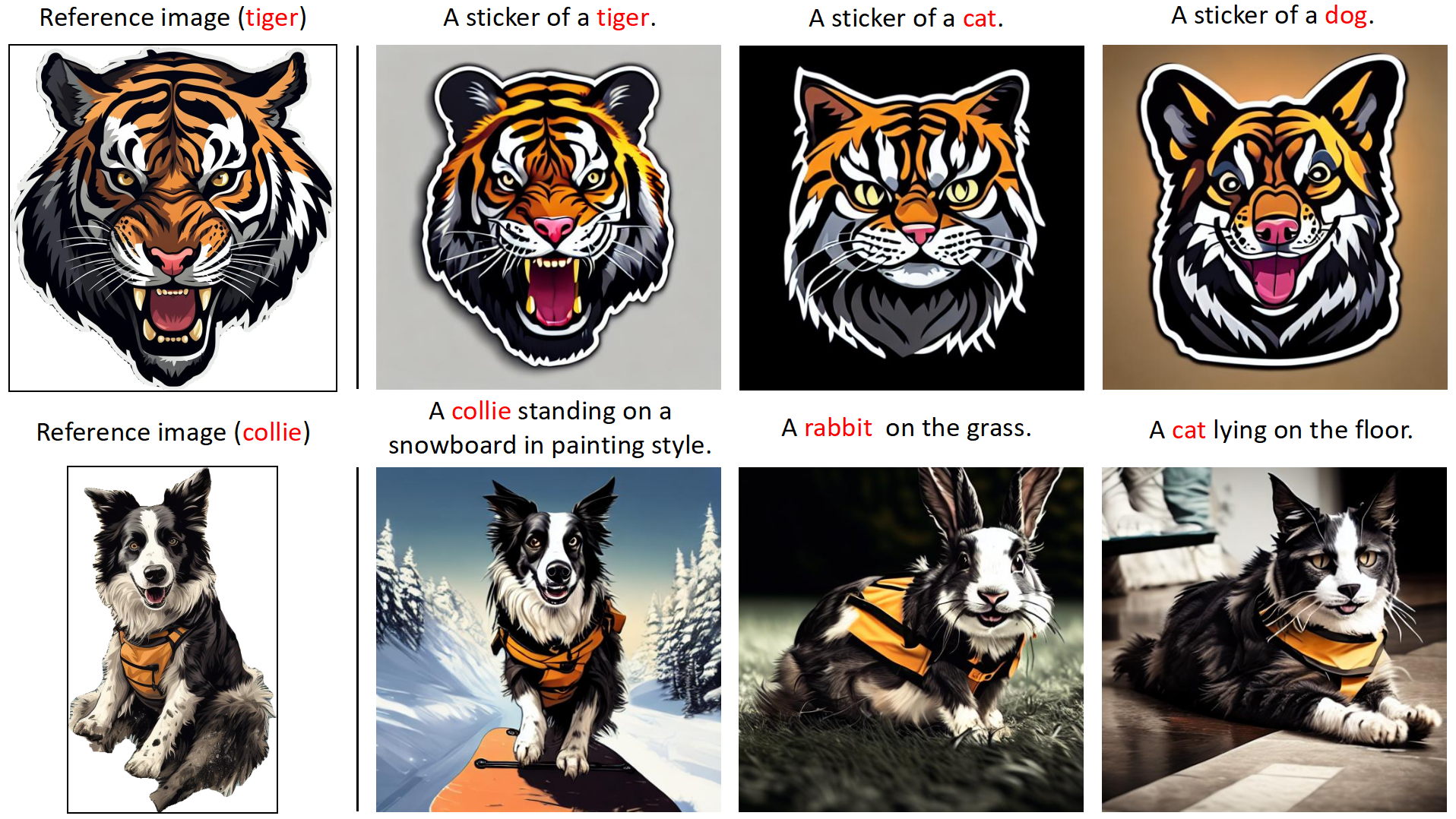}
    \caption{Applications of CustAny on text-image ID mixing, such as merging the text "dog" with the image "tiger". \label{fig:idmixing} }  
    \vspace{-0.15in}
\end{figure}

\noindent \textbf{Text-image ID mixing}.
If the category of the interested object in the text prompt and that in the reference image is not the same, our CustAny can merge the two and form a new ID, as shown in Fig.~\ref{fig:idmixing}

\begin{figure}
    \setlength{\abovecaptionskip}{0cm}
    \centering
    \includegraphics[width=1.0\columnwidth]{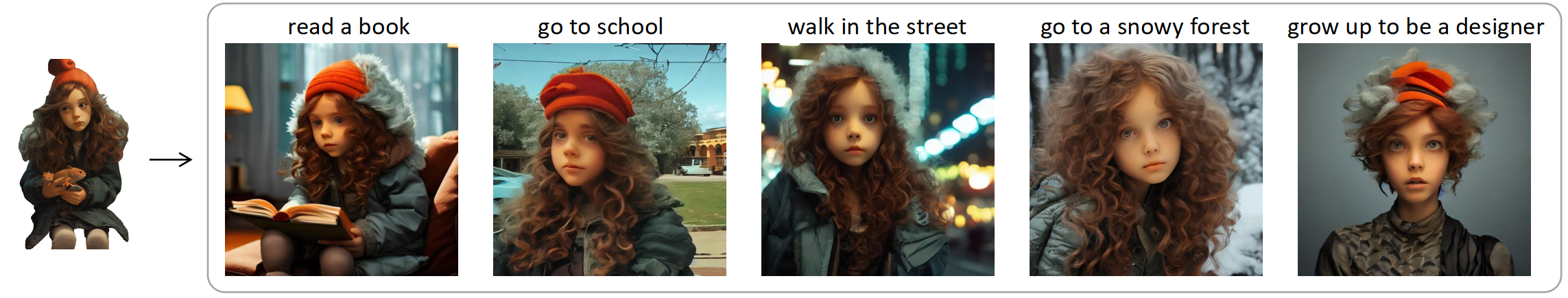}
    \caption{Applications of CustAny on story generation. Given continuous text prompts and one image example, CustAny can generate corresponding stories. \label{fig:story} }  
    \vspace{-0.25in}
\end{figure}

\noindent \textbf{Story generation}.
Our CustAny can generate diverse images under the guidance of text prompts, while maintaining the same identity as the object of interest in the reference image, thereby enabling the creation of a cohesive narrative, as shown in Fig.~\ref{fig:story}.

\section{Conclusion}

We introduce CustAny, a zero-shot text-to-image framework for general object customization, integrating general ID extraction, dual-level ID injection, and an ID-aware decoupling module. CustAny achieves high ID fidelity while preserving text editing abilities and outperforms specialized methods in certain domains. Additionally, we create the MC-IDC dataset, the first large-scale general ID dataset, promote the research for object customization.


\clearpage
\maketitlesupplementary

\section{Additional Details about MC-IDC}
\noindent \textbf{Main categories}. We record several main categories that appear most frequently in MC-IDC, as shown in Tab.~\ref{tab:data_categories}.

\noindent \textbf{Data sources}. The data sources of MC-IDC can be divided into three categories: public datasets, web-crawled images, and movies. We detail the statistical information about various data sources in Tab.~\ref{tab:data_sources}.

\begin{table} \small
\centering
\caption{Sample numbers of main categories in MC-IDC.}
\vspace{-0.1in}
\begin{tabular}{cc} 
\toprule
Categories & Images \\ 
\midrule
man & 46,720 \\
woman & 26,670 \\
clothes & 20,040 \\
girl & 3,498 \\
panda & 3,007 \\
train & 2,286 \\ 
car	& 1,974 \\
boy	& 1,855 \\
dog	& 1,820 \\
\bottomrule
\end{tabular}
\vspace{-0.1in}
\label{tab:data_categories}
\end{table}



\begin{table*} \small
\centering
\caption{Details about data sources of MC-IDC.}
\vspace{-0.1in}
\begin{tabular}{cccc}
\toprule
Source & Dataset & Type & Image pair numbers \\ 
\midrule
\multirow{6}*{Public datasets}
& HumanFace~\cite{bain2021webvid} & Video & 55,830 \\
& VOS~\cite{xu2018youtubevos} & Video & 55,823 \\
& VIPSEG~\cite{miao2022vipseg} & Video & 27,983 \\
& MVImgNet~\cite{yu2023mvimgnet} & Multi-view image & 53,909 \\
& VITON~\cite{han2018viton} & Multi-view image & 20,000 \\
& LVIS~\cite{gupta2019lvis} & Single image & 8,003 \\
\midrule
\multirow{1}*{Web-crawled images}
& - & Single image & 55,829 \\
\midrule
\multirow{1}*{Movies}
& - & Video & 38,405 \\
\bottomrule
\end{tabular}
\label{tab:data_sources}
\end{table*}

\section{Additional Details about Experiment Setup}
\noindent \textbf{Categories in evaluation dataset}. The evaluation dataset can be divided into general objects, human data, and virtual try-on data.
The human data and the virtual try-on data each contain 300 different samples. General objects in the evaluation dataset consist of 50 categories, each of which contains 8 diverse samples. We summarize the 50 categories in Tab.~\ref{tab:eval_data_categories}.

\begin{table*} \small
\centering
\caption{Categories of general objects in the evaluation dataset.}
\vspace{-0.1in}
\begin{tabular}{ccccc} 
\toprule
 Winter melon & Cabbage & Vessel & Pillow &  Screw driver\\
 Pants & Computer mouse & Lipstick & Rice cooker &  Toy figure\\
 Clothing & Pineapple & Can & Plush toy &  Grape\\
 Toilet paper & Paper box & Skirt & Pawpaw &  Ginger\\
 Bowl & Train & Bottle & Cantaloupe &  Sanitary napkin\\
 Soccer & Bag & Umbrella & Hammer &  Book\\
 Flower & Shoe & Towel & Ashcan &  Telephone\\
 Faucet & Flowerpot & Motorcycle & Mug &  Kiwi\\
 Pot & Grapefruit & Jug & Car &  Basket\\
 Balloons & Tomato & Flashlight & Bagged snacks &  Toy duck\\
\bottomrule
\end{tabular}
\label{tab:eval_data_categories}
\end{table*}

\noindent \textbf{Text prompts for calculating DiverSim-i}. We use diverse text prompts describing different scenarios to guide the generation, and calculate DiverSim-i among the generated images. We record the text prompts in Tab.~\ref{tab:eval_data_prompts}.
\begin{table*} \small
\centering
\caption{Text prompts for calculating DiverSim-i.}
\vspace{-0.1in}
\begin{tabular}{cc} 
\toprule
 Scenarios & Text prompts \\
 \midrule
\multirow{2}*{Snow}
& Original text prompt + "The scene of the picture is in the snow." \\
& Original text prompt + "The background of the picture is in the snow." \\
\midrule
\multirow{2}*{Grass}
& Original text prompt + "The scene of the picture is on the grass." \\
& Original text prompt + "The background of the picture is on the grass." \\
\midrule
\multirow{2}*{Beach}
& Original text prompt + "The scene of the picture is on the beach." \\
& Original text prompt + "The background of the picture is on the beach." \\
\midrule
\multirow{2}*{Jungle}
& Original text prompt + "The scene of the picture is in the jungle." \\
& Original text prompt + "The background of the picture is in the jungle." \\
\midrule
\multirow{2}*{Eiffel Tower}
& Original text prompt + "The scene of the picture is beside the Eiffel Tower." \\
& Original text prompt + "The background of the picture is beside the Eiffel Tower." \\
\bottomrule
\end{tabular}
\label{tab:eval_data_prompts}
\end{table*}

\section{More Visual Results}

\begin{figure*}
    \centering
    \includegraphics[width=\linewidth]{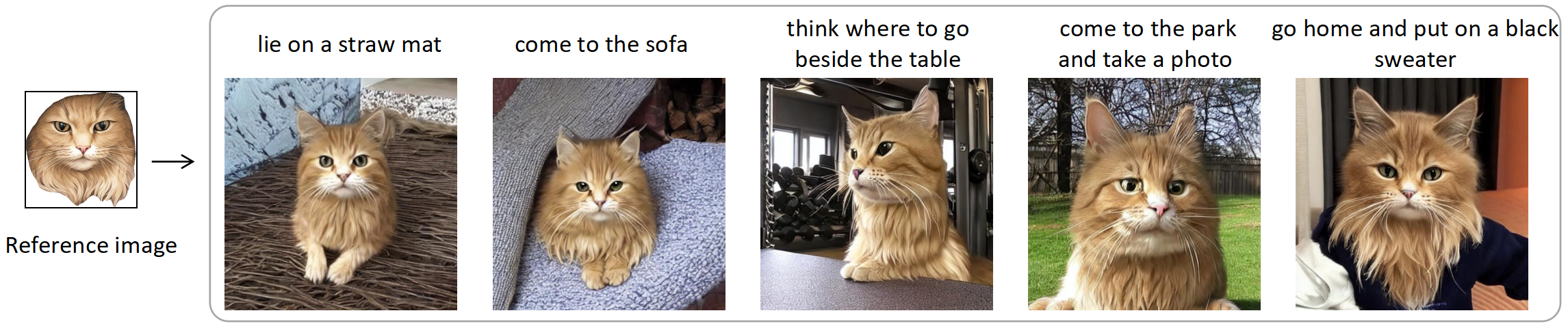}
    \caption{Additional visual results of story generation. Our CustAny can generate diverse images under the guidance of text prompts, while maintaining the same identity as the object of interest in the reference image, thereby enabling the creation of a cohesive narrative.     \label{fig:k2-fulu-story}}
\vspace{-0.2in}
\end{figure*}

\begin{figure*}
    \centering
    \includegraphics[width=\linewidth]{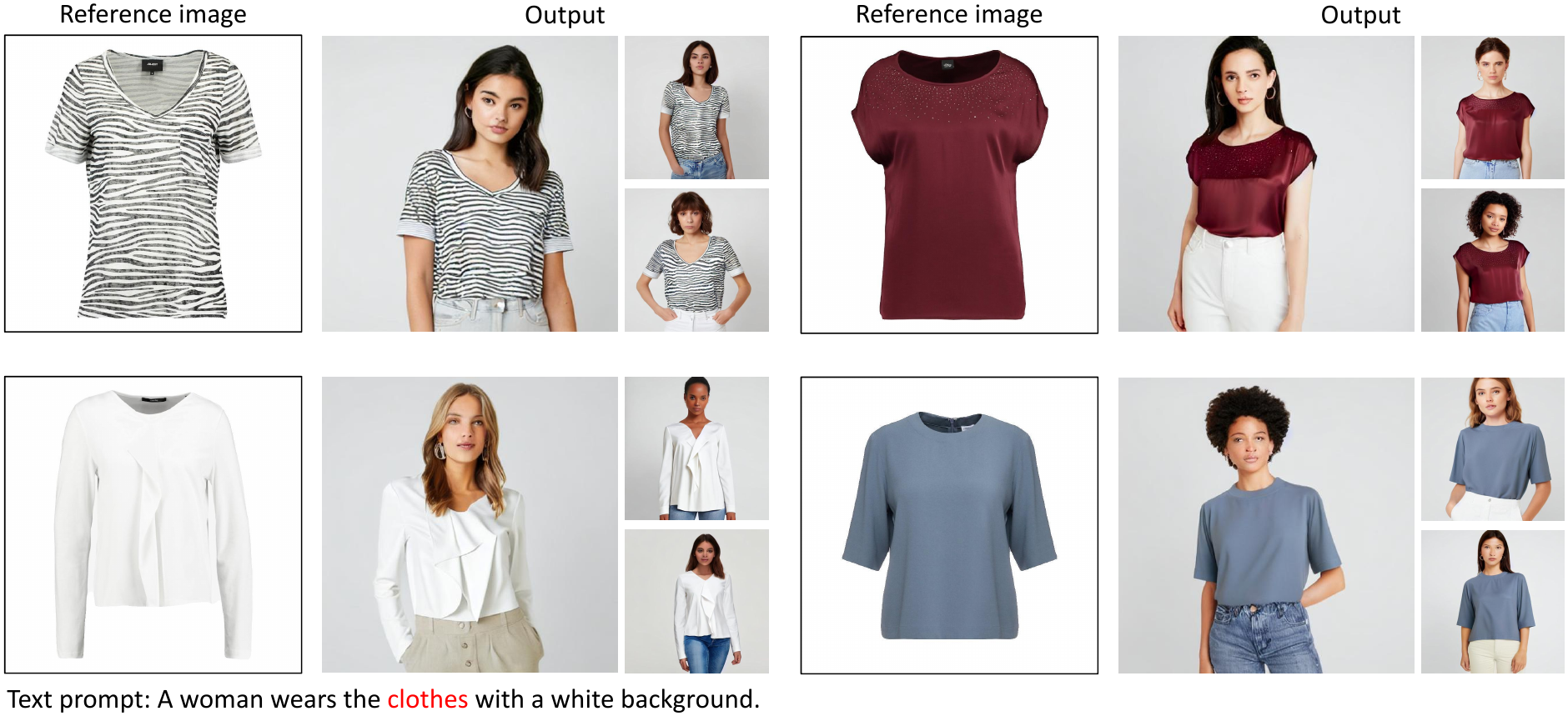}
    \caption{Additional visual results of virtual try-on. Given a piece of clothing, the CustAny can generate images of the clothing worn on a person.     \label{fig:k2-fulu-virtual}}
\vspace{-0.2in}
\end{figure*}

\begin{figure*}
    \centering
    \includegraphics[width=\linewidth]{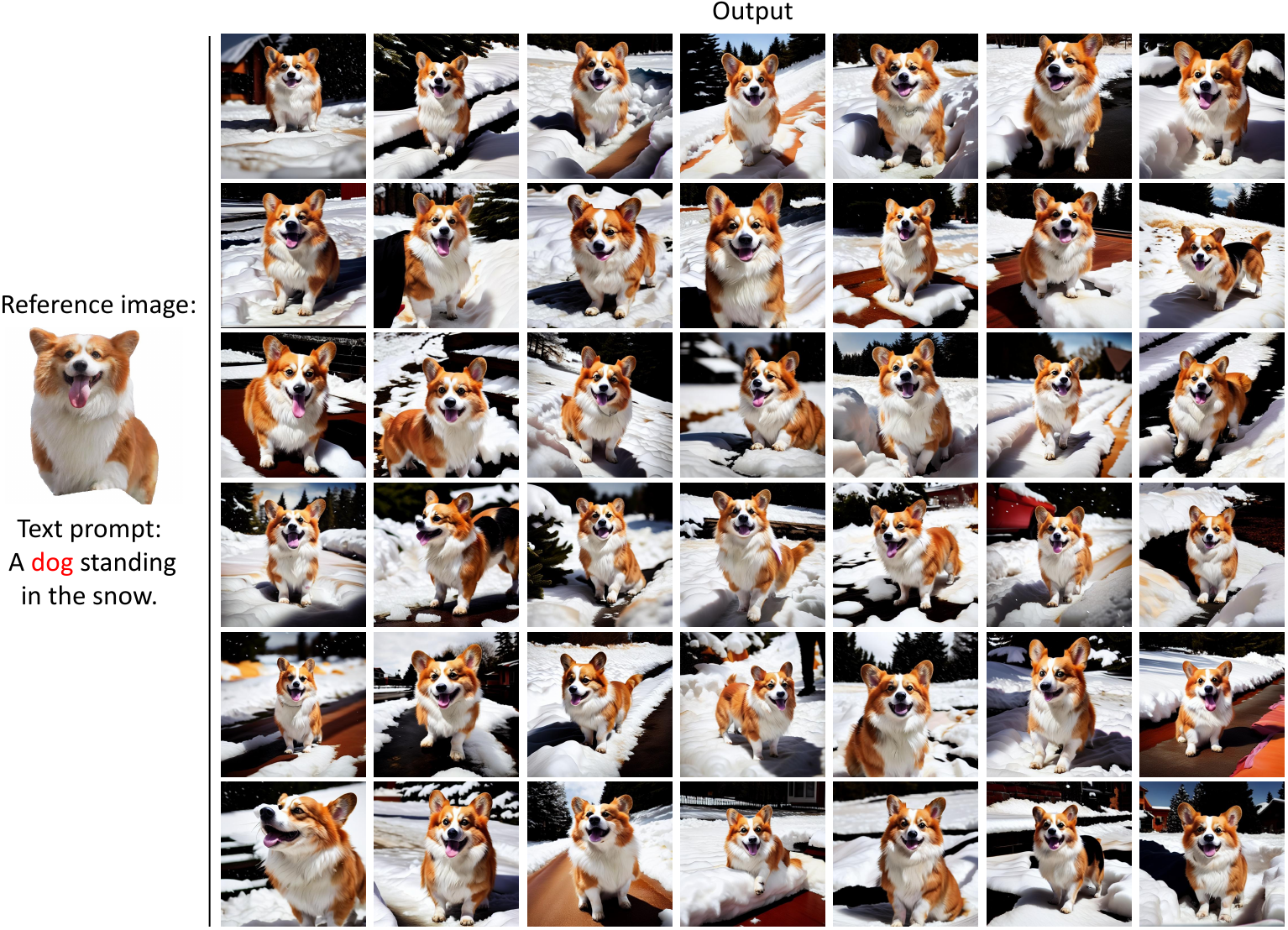}
    \caption{Additional visual results of ID-consistent generation. The CustAny can generate multiple ID-consistent images with diverse non-ID elements such as motions and orientations.     \label{fig:k2-fulu-idconsis}}
\vspace{-0.2in}
\end{figure*}

Our CustAny exhibits outstanding performance on various applications, such as story generation in Fig.~\ref{fig:k2-fulu-story}, virtual try-on in Fig.~\ref{fig:k2-fulu-virtual}, and ID-consistent generation in Fig.~\ref{fig:k2-fulu-idconsis}. 
Our CustAny can ensure both the ID fidelity and the generating diversity simultaneously.

\clearpage
{
    \small
    \bibliographystyle{ieeenat_fullname}
    \bibliography{main}
}


\end{document}